\documentclass[runningheads]{llncs}
\usepackage{graphicx}

\usepackage{tikz}
\usepackage{comment}
\usepackage{amsmath,amssymb} 
\usepackage{color}
\usepackage{booktabs}
\usepackage{verbatim}

\usepackage[accsupp]{axessibility}  

\usepackage[pagebackref,breaklinks,colorlinks]{hyperref}

\usepackage{graphicx,shortbold,xspace,xcolor,colortbl}

\newcommand{\dfnote}[1]{\textcolor{red}{DF: #1}}
\newcommand{\crnote}[1]{\textcolor{blue}{CR: #1}}
\newcommand{\parnobf}[1]{\vspace{0.5mm} \par \noindent {\bf {#1}.}}
\newcommand{\parnoit}[1]{\noindent {\it {#1}.\xspace}}

\definecolor{ibm1}{HTML}{648FFF}
\definecolor{ibm2}{HTML}{DC267F}
\definecolor{ibm3}{HTML}{FE6100}
\definecolor{ibm4}{HTML}{FFB000}
\definecolor{ibm5}{HTML}{785EF0}
\definecolor{ibm6}{HTML}{88CCEE}

\begin{document}
\pagestyle{headings}
\mainmatter
\def\ECCVSubNumber{3429}  

\title{PlaneFormers: From Sparse View Planes to 3D Reconstruction} 


\titlerunning{PlaneFormers}
%

\index{Fouhey, David F.}
\author{Samir Agarwala \and
Linyi Jin \and Chris Rockwell \and David F. Fouhey} 
\authorrunning{Agarwala et al.}
%
\institute{University of Michigan, Ann Arbor\\
\email{\{samirag,jinlinyi,cnris,fouhey\}@umich.edu}}
\maketitle

\begin{abstract}
We present an approach for the planar surface reconstruction of a scene from
images with limited overlap. This reconstruction task is challenging since it requires jointly reasoning about single image 3D reconstruction, correspondence between images, and the relative camera pose between images. Past work has proposed optimization-based approaches. We introduce a simpler approach, the PlaneFormer, that uses a transformer applied to 3D-aware plane tokens to perform 3D reasoning. Our experiments show that our approach is substantially more effective than prior work, and that several 3D-specific design decisions are crucial for its success. Project page: \url{https://samiragarwala.github.io/PlaneFormers}.
\end{abstract}

\begin{figure}[h]
\vspace{-0.5em}
	\centering
			{\includegraphics[width=\linewidth]{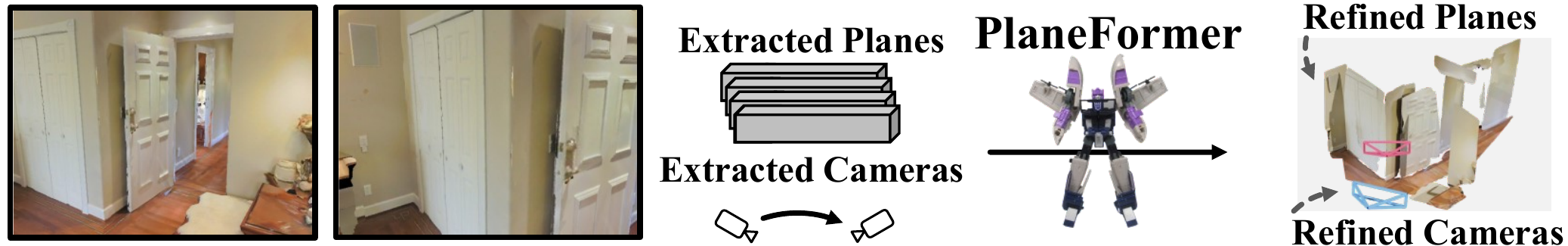}}
\caption{Given a sparse set of images, our method detects planes and cameras, and produces plane correspondences and refined cameras using a Plane Transformer (PlaneFormer~\cite{planeformer}), from which it can reconstruct the scene in 3D.}
	\label{fig:teaser}
\end{figure}

\section{Introduction}
Consider the two images shown in Figure \ref{fig:teaser}. Even though you are not provided with the relative pose between the cameras that took the pictures you see, and even though you have never been to this particular location, you can form a single coherent explanation of the scene. You may notice, for instance, the doors and closet that are visible in both pictures. From here, you can deduce the relative positioning of the cameras and join your 3D perception of each image. The goal of this paper is to further the ability of computers to solve this problem. 

The {\it sparse view} (wide and unknown baseline, few image) setting is challenging for existing systems because it falls between two main strands of 3D reconstruction in contemporary computer vision: multiview 3D reconstruction (usually by correspondence) and learned single view 3D reconstruction (usually by statistical models). In particular, traditional multiview tools \cite{Hartley04,schonberger2016structure,schoenberger2016mvs,Pritchett98a,Jin2020} depend heavily on triangulation as a cue. Thus, in addition to struggling when view overlap is small, their cues usually entirely fail with no overlap. While single-view tools~\cite{hoiem2005geometric,eigen15,liu2019planercnn} can reconstruct single views via learning, merging the overlap between the views to produce one coherent reconstruction is challenging: identifying whether one extracted wall goes with another requires understanding appearance, the local geometry, as well as the relationship between the cameras.

Existing approaches in this multiview area have key limitations in either input requirements or approach. Many approaches assume known camera poses~\cite{kar2017learning,mildenhall2020nerf}, which fundamentally changes the problem by restricting a search for correspondence for a pixel in one image to a single line in another~\cite{Hartley04}. While some works relax the assumption~\cite{lin2021barf} or avoid it via many images~\cite{huang2018deep}, these have not been demonstrated in the few-image, wide baseline case. Most work in the sparse view setting (e.g.,~\cite{Cai2021Extreme,Chen_2021_CVPR}) does pose estimation but not reconstruction and works that produce reconstructions from sparse views~\cite{Qian2020,Jin2021} come with substantial limitations. Qian et.~al~\cite{Qian2020} require multiple networks,  watertight synthetic ground-truth, and use a heuristic RANSAC-like search. Jin et.~al~\cite{Jin2021} apply a complex hand-designed discrete/continuous optimization applied to plane segments found by an extended PlaneRCNN~\cite{liu2019planercnn} output. This optimization includes bundle-adjustment on SIFT~\cite{lowe2004distinctive} on viewpoint-normalized texture like VIP~\cite{wu20083d}.

We propose an approach (\S\ref{sec:approach}), named the PlaneFormer, that overcomes these limitations. Following existing work in this area~\cite{Qian2020,Jin2021}, we construct a scene reconstruction by merging scene elements that are visible in multiple views and estimating relative camera transformations. We build on~\cite{Jin2021} and construct a piecewise planar reconstruction from the images. However, rather than perform an optimization, we directly train a transformer that ingests the scene components as tokens. These tokens integrate 3D knowledge and a working hypothesis about the relative pose between input views. As output, this transformer estimates plane correspondence, predicts the accuracy of the working hypothesis for the relative poses, as well as a correction to the poses. By casting the problem via transformers, we eliminate manual design and tuning of an optimization. Moreover, once planes are predicted, our reconstruction operations are performed via transformer forward passes that test out hypothesized relative camera poses.

Our experiments (\S\ref{sec:experiments}) on Matterport3D~\cite{chang2017matterport3d} demonstrate the effectiveness of our approach compared to other approaches. We evaluate with set of image pairs with limited overlap (mean rotation: $53^\circ$; translation: $2.3$m; overlap: 21\%). Our approach substantially surpasses the state of the art~\cite{Jin2021} before its post-processing bundle adjustment step: the number of pairs registered within 1m increases from 56.5\% to 66.8\%, and pairs with 90\% correspondences correct increases from 28.1\% to 40.6\%. Even when~\cite{Jin2021} uses the additional bundle adjustment step, the our approach matches or exceeds the method. We next show that our approach can be used on multiple views, and that several 3D design decisions in the construction of the PlaneFormer are critical to its success.

\section{Related Work}







Our approach to 3D reconstruction from sparse views draws upon the well-studied tasks of correspondence estimation, i.e., 3D from many images; and learning strong 3D priors, i.e., 3D from a single image.

\parnobf{Correspondence and camera pose estimation} The tasks of estimating correspondences and relative camera pose~\cite{banani2020unsupervisedrr,sarlin2020superglue,choy2020deep,zhang1994iterative,yi2018learning,zhang2019learning} across images are central to predicting 3D structure from multiple images~\cite{bozic2021transformerfusion,schonberger2016structure,huang2018deepmvs,sun2021neuralrecon}. Some methods jointly refine camera and depth across many images~\cite{lindenberger2021pixel,teed2021droid,mur2015orb,lin2021barf,zhang2021consistent} in a process classically approached via Bundle Adjustment~\cite{triggs1999bundle,agarwal2010bundle}. We also refine both camera and reconstruction; however, we do not have the requirement of many views. Additionally, we use self-attention, a powerful concept that has been successfully used in several vision tasks \cite{sarlin2020superglue,Sun2021,lin2021metro,bozic2021transformerfusion,zhao2021point}. Our approach of using self-attention through transformers \cite{vaswani2017attention}  is  similar to SuperGlue~\cite{sarlin2020superglue} and LoFTR~\cite{Sun2021} in that it permits joint reasoning over the set of potential correspondences. We apply it to the task of planes, and also show that the learned networks can also predict relative camera pose directly (via residuals to a working hypothesis).

\parnobf{3D from a single image} Learned methods have enabled 3D inference given only a single viewpoint. These methods cannot rely on correspondences, and therefore use image cues along with learned priors and a variety of representations. Their 3D structure representations include voxels~\cite{choy20163d,song2017semantic}, meshes~\cite{gkioxari2019mesh,wang2018pixel2mesh}, point clouds~\cite{fan2017point,wiles2020synsin}, implicit functions~\cite{mescheder2019occupancy,jiang2020local}, depth~\cite{Ranftl2020,li2018megadepth}, surface normals~\cite{Wang15,chen2020oasis}, and planes~\cite{liu2018planenet,yang2018recovering,yu2019single}. We use planes to reconstruct 3D as they are often good approximations~\cite{furukawa2009manhattan} and have strong baselines for detection such as PlaneRCNN~\cite{liu2019planercnn}; we build off this architecture. In contrast with PlaneRCNN and single-image methods, we incorporate information across multiple views and therefore can also use correspondences. 

\parnobf{3D from sparse views} Recent approaches enable learned reasoning with multiple views. Several works perform novel view synthesis using radiance fields \cite{yu2020pixelnerf,jain2021putting,wang2021ibrnet,chen2021mvsnerf}. Learned methods also estimate pose~\cite{Cai2021Extreme,wang2020tartanvo} and depth~\cite{ummenhofer2017demon,kopf2021robust} given few views, but do not create a unified scene reconstruction. Our focus is also on wide-baseline views~\cite{Pritchett98a}, further separating us from monocular stereo methods~\cite{ummenhofer2017demon,kopf2021robust}. Two recent works approach this task. Qian \textit{et al.}~\cite{Qian2020} reconstruct objects from two views but use heuristic stitching across views, and struggle on realistic data~\cite{Jin2021}. Jin \textit{et al.}~\cite{Jin2021} jointly optimize plane correspondences and camera pose from two views with a hand-designed optimization. In contrast, we use a transformer to directly predict plane correspondence and camera pose. Our experiments (\S\ref{sec:experiments}) show our approach outperforms these methods. 

\section{Approach}
\label{sec:approach}

Our approach aims to jointly reason about a pair of images with an unknown relationship and reconstruct a single, coherent, global planar reconstruction of the scene depicted by the images. This process entails extracting three key related pieces of information: the position of the planes that constitute the scene; the correspondence between the planes in each view so that each real piece of the scene is reconstructed once and only once; and finally, the previously unknown relationship between the cameras that took each image.

At the heart of our approach is a plane transformer that accepts an initial independent reconstruction of each view and hypothesized global coordinate frame for the cameras. In a single forward pass, the plane transformer identifies which planes correspond with each other, predicts whether the cameras have  correct relative pose, and estimates an updated relative camera pose as a residual. Inference for the scene consists of running one forward pass of the PlaneFormer network per camera hypothesis. 

\subsection{Backbone Plane Predictor}
\label{sec:backbone}

The PlaneFormer is built on top of a single-view plane estimation backbone from \cite{Jin2021}, which is an extended version of PlaneRCNN~\cite{liu2019planercnn}. We refer the reader to~\cite{Jin2021} for training details, but summarize the key properties here. This plane backbone produces two outputs: per-image planes and a probability distribution over relative camera poses. 

\parnoit{Plane Branch} 
The per-image planes are extracted from an image $I_i$ via a Mask-RCNN~\cite{he2017mask}-like architecture. This architecture detects a set of plane segments, yielding $M_i$ detections. Each detected segment in the $i$th image is indexed by $j$ and has a mask segment $\mathcal{S}_{i,j}$, plane parameters $\piB_{i,j} \in \mathbb{R}^4$, and appearance embedding $\eB_{i,j} \in \mathbb{R}^{128}$. The plane parameter $\piB_{i,j}$ can further be factored into a normal $\nB_{i,j}$ and offset $o_{i,j}$ (defining a plane equation $\nB_{i,j}^T [x,y,z] - o_{i,j} = 0$). The appearance embedding can be used to match between images $i$ and $i'$: the distance in embedding space $||\eB_{i,j}-\eB_{i',j'}||$ ought to be small whenever plane $j'$ corresponds to the plane $j$. 

\parnoit{Camera Branch} 
The backbone also produces a probability distribution over a predefined codebook of relative camera transformations $\{\hat{\RB}_k, \hat{\tB}_k\}$.
To predict this distribution, it uses a CNN applied to cross-attention features between early layers of the
network backbone (specifically, the 
{\tt P3} layers of the ResNet-50-FPN~\cite{lin2017feature}). This camera branch combines cross-attention features~\cite{Jin2021,Cai2021Extreme} and pose via regression-by-classification~\cite{Qian2020,Jin2021,Chen_2021_CVPR,Cai2021Extreme}, both of which have been shown to lead to strong performance. While a strong baseline, past work~\cite{Qian2020,Jin2021} has shown that the predictions of such networks need to be coupled with reasoning. For instance,~\cite{Jin2021} uses the probability for each camera pose as a term in its optimization problem. In our case, we use it to generate a set of initial hypotheses about the relative camera poses.


\subsection{The PlaneFormer}

The core of our method is a transformer~\cite{vaswani2017attention,bloem_2019} that jointly processes the planes detected in the 2 images given a hypothesized global coordinate system for the images. Since transformers operate on sets of inputs using a self-attention mechanism, they are able to consider context from all inputs while making predictions. This makes them effective in tasks such as ours where we want to collectively reason about multiple planes across the images to generate a coherent reconstruction. In our case, the transformer takes in a set of feature vectors representing plane detections as the input and maps them to an equal number of outputs which are then further processed and passed through MLP heads to predict our outputs. This function can take in a variable number of feature vectors as an input and is learned end-to-end. The plane transformer aims to identify:  correspondence between the planes (i.e., whether they represent the same plane in 3D and can be merged), whether the hypothesized relative camera pose is correct, and how to improve the relative camera pose.

\begin{figure}[t!]
	\centering
			{\includegraphics[width=\linewidth]{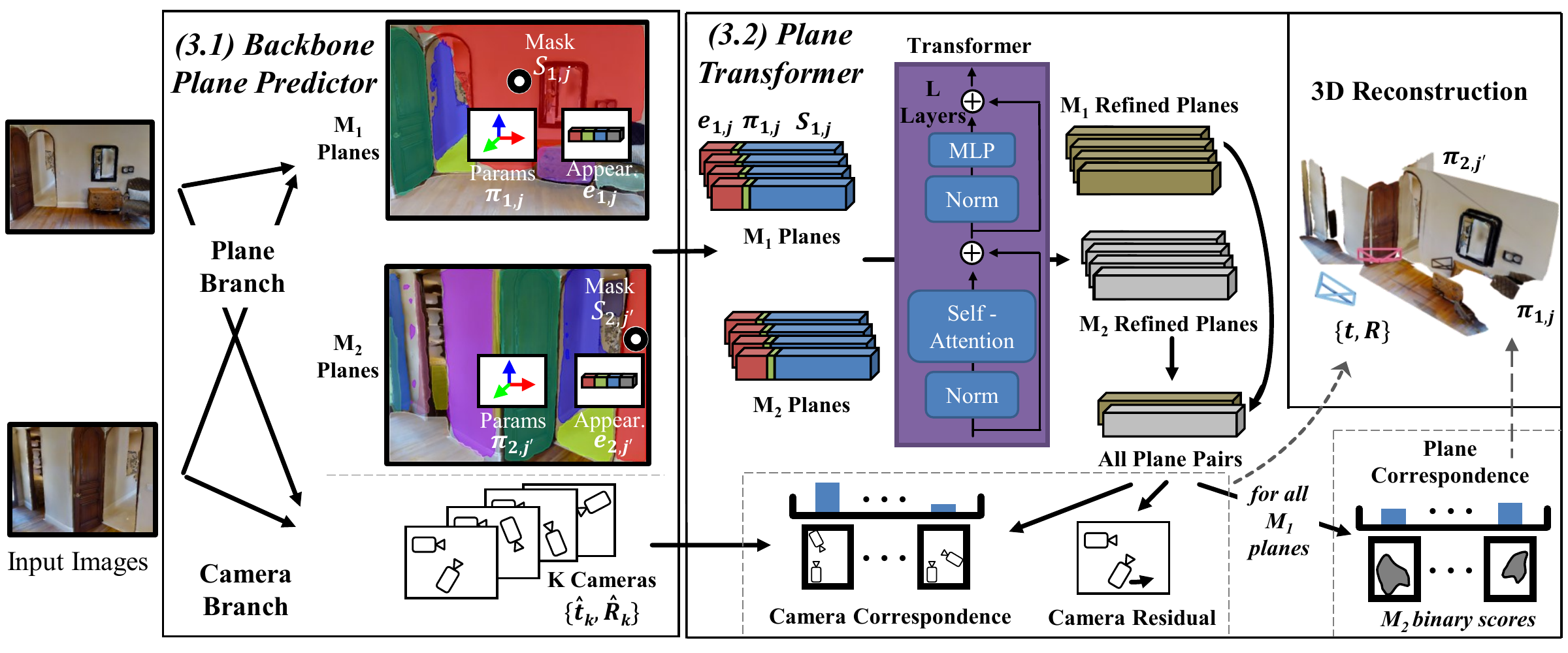}}
\caption{\textbf{Approach.} Given two input images, the backbone network detects planes and predicts camera pose across images. The plane transformer refines these planes by predicting correspondence and refined camera pose, producing a final 3D reconstruction.}
	\label{fig:method}
\end{figure}

As input, the plane transformer takes $M = M_1 + M_2$ tokens. These tokens contain features that integrate the hypothesized coordinate frame to help the transformer. The hypothesized coordinate frame consists of rotations and translations $\{\RB_i,\tB_i\}$ that are hypothesized to bring the two camera views into a common coordinate frame. For convenience, we assume that the common coordinate frame is centered at image $2$ (i.e., $\RB_2 = \IB_{3\times3}$, $\tB_2 = \zeroB$), but note that the transformer never has explicit access to these rotation and translations. 

\parnobf{Input Features} We concatenate three inputs to the network from the plane backbone to represent each of the $M$ plane tokens. Each token is 899D.

\parnoit{Appearance Features (128D)} The first token part is the appearance feature from \cite{Jin2021}'s extended PlaneRCNN.

\parnoit{Plane Features (3D)} The second part of the token is the plane equation $\piB_{i,j}$ comprising a normal that is scaled by the plane offset. This equation is transformed to the hypothesized canonical space by $\RB_{i}, \tB_i$. This feature functions like a positional encoding, and enables logic such as: if two planes have a similar appearance and plane parameters, then they are likely the same plane. 

\parnoit{Mask Features (768D)} We directly provide information about the plane segments via mask features. These features complement the plane features since they represent a plane segment rather than an infinite plane. We use the hypothesized relative camera pose and 3D to produce mask features. The mapping from image $i$ to the common coordinate frame's view for plane $j$ is given by a homography $\HB = \RB_i + (\tB_i^T \nB_{i,j})/o_{i,j}$~\cite{ma2004invitation}. This lets us warp each mask $\mathcal{S}_{i,j}$ to a common reference frame. Once the mask is warped to the common reference frame, we downsample it to a $24 \times 32$ image. We hypothesize that the explicit representation facilitates reasoning such as: these two planes look the same, but they are on opposite sides so the provided transformation may be wrong; or these planes are the same and roughly in the right location but one is bigger, so the translation ought to be adjusted. 

\parnobf{Outputs} As output, we produce a set of  tensors that represent plane correspondence, whether cameras have the correct relative transformation, and updated camera transformations. Specifically, the outputs are:

\parnoit{Plane Correspondence} $\PiB \in \mathbb{R}^{M_1\times M_2}$ that gives the correspondence score between two planes across the input images. If $\PiB_{j,j'}$ is large, then the planes $j$ and $j'$ likely are the same plane in a different view. We minimize a binary cross-entropy loss between the predicted $\PiB$ and ground-truth..

\parnoit{Camera Correspondence} $C \in \mathbb{R}$ that indicates whether the two cameras have the correct relative transformation. If $C$ has a high score, then it is likely that the hypothesized relative relative pose between the input cameras is correct. We minimize a binary cross-entropy loss between the predicted $C$ and ground-truth.

\parnoit{Camera Residual} $\DeltaB \in \mathbb{R}^{7}$ giving a residual for the hypothesized relative pose. This residual is expressed as the concatenation of a 4D quaternion for rotation and 3D translation vector for translation. Updating the relative transformation between the cameras is likely to improve the transformation. We minimize an $L_1$ loss between the predicted camera rotation and translation residual and the ground-truth camera rotation and translation residual with relative weight $\lambda_t$ to translation. During training, hypothesized camera poses come from the codebook from the Camera Branch; thus there is a residual that needs to be corrected.

\parnobf{PlaneFormer Model} A full description of the method appears in the supplement, but our PlaneFormer consists of a standard transformer followed by the construction of pairwise features between planes. The transformer maps the $M$ plane input tokens to $M$ output tokens using a standard Transformer~\cite{vaswani2017attention} with 5 layers and a feature size equal to plane tokens of $899$. We use only a single head to facilitate joint modeling of all plane features.

After the transformer produces $M$ per-plane output tokens, the $M$ outputs are expanded to $M \times M$ pairwise features in an outer-product-like fashion. To assist in prediction, we also produce a per-image token: given $M$ output tokens, where $\oB_{i,j}$ denotes the output token for the $j$th plane in image $i$, we compute the per-image average token $\muB_i = (1/M_i) \sum_{j=1}^{M_i} \oB_{i,j}$. The pairwise feature for planes $(i,j)$ and $(i',j')$ is the concatenation of the plane output tokens $\oB_{i,j}$, $\oB_{i',j'}$, and their per-image tokens $\muB_{i}$, and $\muB_{i'}$. This 3596 ($4 \times 899$) feature is passed into separate 4-hidden-layer MLP heads that estimate $\PiB$, $C$, and $\DeltaB$ per-pair of planes. At each hidden layer, we halve the input feature dimension. We average pool the MLP output over plane pairs across the images to produce the final estimate for $C$ and $\DeltaB$. $\PiB$ can be used after masking to $M_1 \times M_2$. Finally, we apply a sigmoid function to $C$ and $\PiB$ to generate the model output.



\subsection{Inference}
\label{sec:method_inference}

Once the PlaneFormer has been trained, we can apply it to solve reconstruction tasks. Given a set of planes and hypothesized poses of cameras, the Planeformer can estimate correspondence, identify whether the hypothesized poses are correct, and estimate a correction to the camera poses. 

\begin{figure}[t!]
	\centering
			{\includegraphics[width=\linewidth]{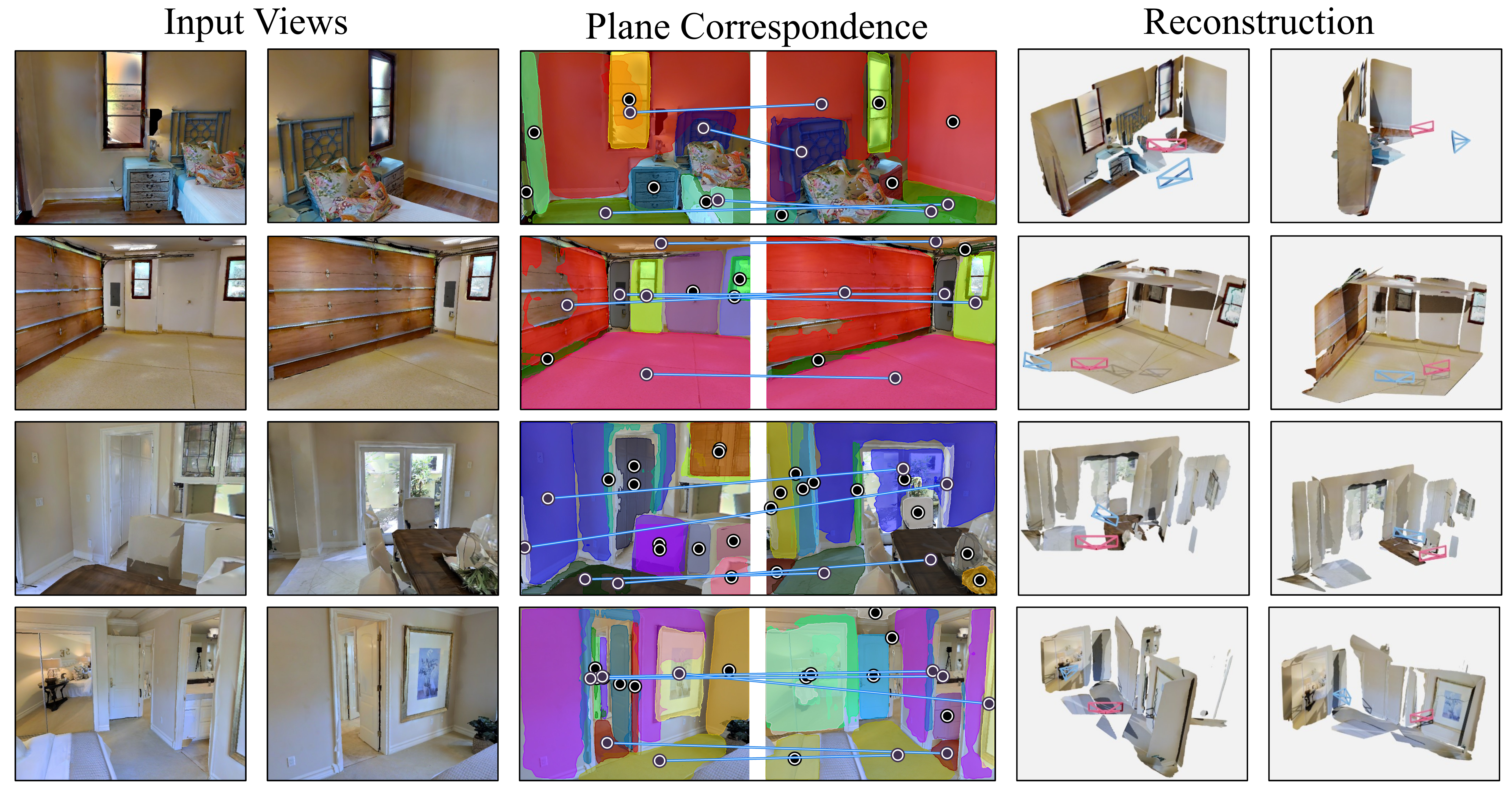}}
\caption{\textbf{Sample Outputs on the Test Set.} PlaneFormer produces jointly refined plane correspondences and cameras, from which it reconstructs the input scene. It can produce high-quality reconstructions in cases of moderate view change (top two rows), and coherent reconstructions in cases of large view change (bottom two rows).}
	\label{fig:results}
\end{figure}

\parnobf{Two View Inference} Given two images, one takes the top $h$ hypotheses for the relative camera pose from the Camera Branch (\S\ref{sec:backbone}) and evaluates them with the PlaneFormer. The pose hypothesis with highest camera correspondence score is selected, and the predicted residual is added. We note that these camera pose hypotheses can be explored in parallel, since they only change the token features. After the plane correspondences have been predicted, we match using the Hungarian Algorithm \cite{kuhn1955hungarian} with thresholding. Sample outputs of PlaneFormer appear in Fig.~\ref{fig:results} and throughout the paper.

\parnobf{Multiview Inference} In order to extend the method to multiple views, we can apply it pairwise to the images. We apply the above approach pairwise to edges in an acyclic view graph that connects the images. The graph is generated greedily on a visibility score for a pair of images $(i,i')$ that represents the number of planes with close matches. For the appearance embedding $\eB_{i,j}$ of plane $j$ in image $i$, we compute the minimum distance to the appearance embeddings of the planes in image $i'$, or  $d_{j} = \min_{j'} ||\eB_{i,j} - \eB_{i',j'}||$. Rather than threshold the distance, we softly count the numbers of close correspondence via a score $\sum_j \exp(-d_j^2 / \sigma^2)$. We repeat the process from $i'$ to $i$ and then sum for symmetry. 


\subsection{Training and Implementation Details}

\parnoit{Training procedure} During training, each sample must assume a set of camera transformations (i.e., $\{\RB_i, \tB_i\}$). We train on a mix of correct camera transformations (using the nearest rotation and translation in the sparse codebook) and incorrect camera transformations (using a randomly selected non-nearest rotation and translation in the sparse codebook). Given a correct camera hypothesis, we backpropagate losses on all outputs; given an incorrect hypothesis, we backpropagate losses only on the camera correspondence $C$. Thus, the camera correspondence output is trained to discriminate between correct and incorrect cameras, and the other outputs do not have their training contaminated (e.g., by having to predict residuals even if the camera hypothesis is completely incorrect). The correct and incorrect cameras are sampled equally during training.

\parnoit{Implementation Details} We train for 40k iterations using a batch size of 40 and the same Matterport3D~\cite{chang2017matterport3d} setup as Jin \textit{et al.}~\cite{Jin2021} We use SGD with momentum of 0.9 and a learning rate of 1e-2, and follow a one-cycle cosine annealing schedule. We weight all losses equally, with the exception of $\lambda_t=0.5$ for the residual translation loss. Training takes about 36 hours using 4 RTX 2080 Tis. At inference, we select from $h=9$ camera hypotheses.

\section{Experiments}
\label{sec:experiments}

We now evaluate the proposed approach in multiple settings. We first introduce our experimental setup, including metrics and datasets. We then introduce experiments for the wide baseline two view case in \S\ref{sec:experiments_two}. The two view setting has an abundance of baselines that we compare with. Next, we introduce experiments for more views, specifically 3 and 5 views, in \S\ref{sec:experiments_multi}. Finally, we analyze which parts of our method are most important in \S\ref{sec:experiments_ablation}.

\parnobf{Metrics} The sparse view reconstruction problem integrates several challenging, complex problems: detecting 3D planes from a 2D image, establishing correspondence across images, and estimating relative camera pose. We therefore evaluate the problem in multiple parts.

\parnoit{Plane Correspondence} We evaluate correspondence separately. We follow Cai et al.~\cite{cai2020messytable} and use IPAA-X, or the fraction of image pairs with no less than X\% of planes associated correctly. We use ground-truth plane boxes in this setting since otherwise this metric measures both plane detection and plane correspondence.

\parnoit{Relative Camera Pose} We next evaluate camera relative pose estimation. We follow~\cite{Qian2020,Jin2021} and report the mean error, the median error, and the fraction of image pairs with error below a threshold of $30^\circ$ and $1$m following~\cite{Qian2020}.

\parnoit{Full Scene Results} Finally, we report results using the full scene metric from~\cite{Jin2021}. This metric counts detected planes as true positives if their mask IoU is $\ge 0.5$, surface normal distance is $\le 30^\circ$; and offset distance is less than $1$m. This metric integrates all three components: to get the planes correct, one needs to reconstruct them in 3D, estimate the relative camera configuration to map the second view's planes into the first view, and identify duplicated planes to suppress false positives. While this metric is important, any component can limit performance, including components we do not alter, like plane detection.

\begin{figure}[t!]
	\centering
			{\includegraphics[width=\linewidth]{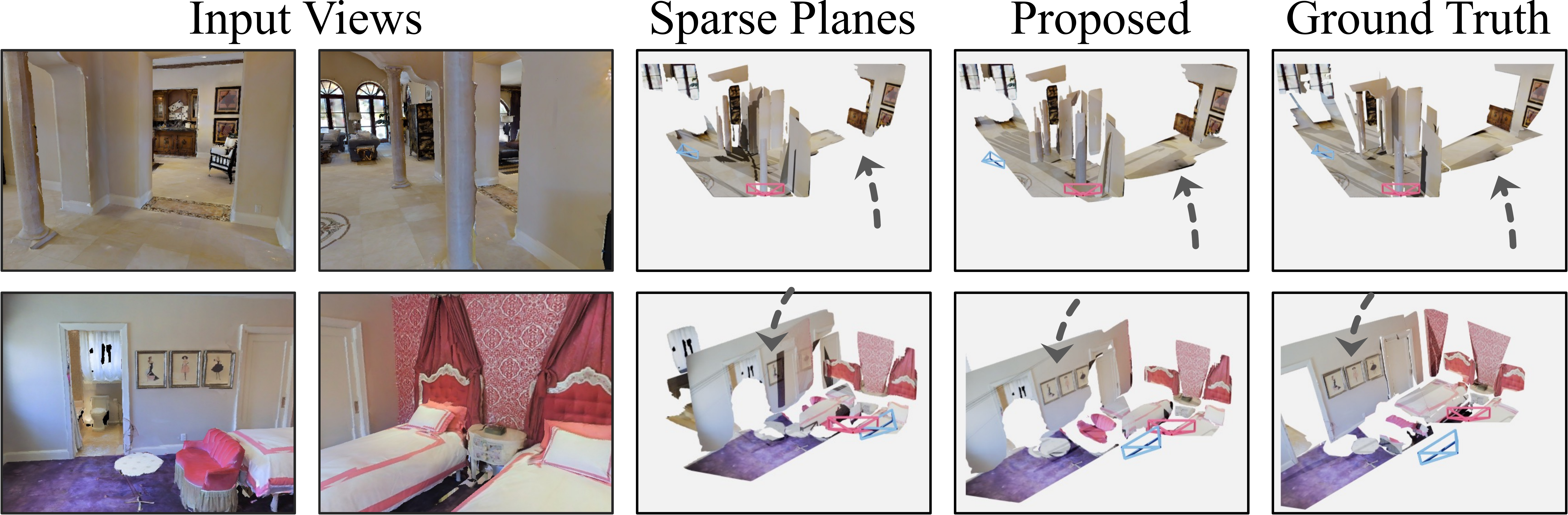}}
\caption{\textbf{Reconstruction Comparison.} Sparse Plane reconstructions are a good, but PlaneFormer's are better in terms of stitched planes (top), and camera poses (bottom).}
	\label{fig:recon_comparison}
\end{figure}

\parnobf{Datasets} We evaluate on three datasets: two-, three-, and five-view.
{\it Two view dataset:} We use the exact dataset used in~\cite{Jin2021} for fair comparison. This consists of 31392 training image pairs, 4707 validation image pairs, and 7996 test image pairs. These views are widely spearated. On average: view overlap is~21\% of pixels; relative rotation is $53^\circ$; and relative translation is $2.3$m. 
{\it Multiview datasets:} We generate a set of $3$- and $5$- view pairs using the same procedure as~\cite{Jin2021}. We evaluate on a total of 258 3-view and 76 5-view samples.

\parnobf{Baselines} The full problem of reconstructing the scene from a set of sparse views requires solving the three separate problems of correspondence, relative camera pose estimation, and 3D reconstruction. We compare with full systems as well as approaches that solve each problem independently.

\parnoit{All Settings} In all cases, we compare with {\it Sparse Planes}~\cite{Jin2021}. For fair comparison, we use an identical backbone to~\cite{Jin2021} so that any performance gain stems from the PlaneFormer rather than improved systems tuning. The full version is our strongest baseline. It uses the same plane information and follows it with a discrete-continuous optimization. The continuous optimization requires extracting view-normalized texture maps, performing SIFT matching, and then bundle adjustment and is expensive and a complementary contribution. Since our method does not do an additional step of extracting feature correspondences and optimizing, a more comparable system to the contribution of our paper is the discrete-optimization only version, or {\it Sparse Planes}~\cite{Jin2021} {\it (No Continuous)}, which performs all the steps except bundle adjustment on point correspondences.

\parnoit{Plane Correspondence} We additionally compare with ({\it Appearance Only}), or the Hungarian algorithm with thresholding on the appearance embedding distances. This approach outperformed other methods like~\cite{cai2020messytable} and \cite{Qian2020} in~\cite{Jin2021}.

\parnoit{Relative Camera Pose Estimation} 
We also compare with a number of other methods. The most important is the ({\it Sparse Planes}~\cite{Jin2021} Camera Branch) which is the top prediction from the Camera Branch network that our system uses for hypotheses. Gain over this is attributable to the PlaneFormer camera correspondence and residual branch, since these produce different relative camera poses. Other methods include: (Odometry~\cite{raposo2013plane} + GT/\cite{Ranftl2020}), which combines a RGBD odometry with ground-truth or estimated depth; ({\it Assoc. 3D}~\cite{Qian2020}), a previous approach for camera pose estimation; ({\it Dense Correlation Volumes}~\cite{Cai2021Extreme}) which uses correlation volumes to predict rotation;  {\it SuperGlue}~\cite{sarlin2020superglue} and {\it LoFTR}~\cite{Sun2021}, which are learned feature matching system. Since~\cite{sarlin2020superglue} and ~\cite{Sun2021} solve for an essential matrix, their estimate of translation is intrinsically scale-free~\cite{Hartley04}. 

\parnoit{Full Scene Reconstruction} 
For full-evaluation, we report some of the top performing baselines from~\cite{Jin2021} along with \cite{Sun2021}. These are constructed by joining the outputs of \cite{Jin2021}'s extended PlaneRCNN~\cite{liu2019planercnn} with a relative camera pose estimation method that gives a joint coordinate frame. These are as described in the relative camera pose estimation, except {\it SuperGlue GT Scale} and {\it LoFTR GT Scale} are also given the ground-truth translation scale. This extra information is needed since the method intrinsically cannot provide a translation scale.

\subsection{Wide Baseline Two-View Case}
\label{sec:experiments_two}


\begin{table}[t]    
    \centering
    \caption{{\bf Two View Plane Correspondence.} IPAA-X~\cite{cai2020messytable} measures the fraction
    of pairs with no less than X\% of planes associated correctly. Ground truth bounding boxes are used. Since the Sparse Planes continuous optimization does not update correspondence, there is not a separate entry for Sparse Planes without continuous optimization.}
    \label{tab:correspondence}
        \begin{tabular}{@{\hskip5pt}l@{\hskip5pt}c@{\hskip5pt}c@{\hskip5pt}c@{\hskip5pt}c@{\hskip5pt}}
        \toprule
        & IPAA-100 & IPAA-90 & IPAA-80 \\
        \midrule
        Appearance Only & 6.8 & 23.5 & 55.7 \\
        Sparse Planes \cite{Jin2021} & 16.2 & 28.1 & 55.3 \\
        Proposed & \textbf{19.6} & \textbf{40.6} & \textbf{71.0} \\
        \bottomrule
        \end{tabular}

\end{table}

\begin{table}[t]    
    \centering
    \caption{{\bf Two View Relative Camera Pose}. We report median, mean error and $\%$ error {$\leq$} 1m or $30^{\circ}$ for translation and rotation.
    }
    \label{tab:camera}
        \begin{tabular}{@{\hskip5pt}l@{~}c@{~}c@{~}c@{~~~~~~}c@{~}c@{~}c@{\hskip5pt}}
        \toprule
        & \multicolumn{3}{c}{Translation} & \multicolumn{3}{c}{Rotation} \\
        Method & Med. & Mean & (${\leq}1$m) & Med. & Mean & (${\leq}30^{\circ}$) \\
        \midrule
        Odometry \cite{raposo2013plane} + GT Depth & 3.20 & 3.87 & 16.0 & 50.43 & 55.10 & 40.9 \\
        Odometry \cite{raposo2013plane} + \cite{Ranftl2020} & 3.34 & 4.00 & 8.3 & 50.98 & 57.92 & 29.9 \\
        Assoc. 3D~\cite{Qian2020} & 2.17 & 2.50 & 14.8  & 42.09 & 52.97 & 38.1 \\
        Dense Correlation Volumes~\cite{Cai2021Extreme} & - & - & - & 28.01 & 41.56 & 52.45 \\
        Camera Branch~\cite{Jin2021} & 0.90 & 1.40 & 55.5 &  7.65 & 24.57 & 81.9 \\
        Sparse Planes \cite{Jin2021} (No Continuous) & 0.88 & 1.36 & 56.5 & 7.58 & 22.84 & 83.7 \\
         Proposed & 0.66 & \textbf{1.19} & \textbf{66.8}  & 5.96 & 22.20 & 83.8 \\   \midrule
         Sparse Planes \cite{Jin2021} (Full) & \textbf{0.63} & 1.25 & 66.6  & 7.33 & 22.78 & 83.4 \\          
        SuperGlue~\cite{sarlin2020superglue} & - & - & - & 3.88 & 24.17 & 77.8 \\
        LoFTR-DS~\cite{Sun2021} & - & - & - & \textbf{0.71} & \textbf{11.11} & \textbf{90.47} \\
        \bottomrule
        \end{tabular}
\end{table}

\begin{table}[t]
    \centering
    \caption{{\bf Two View Evaluation.} Average Precision, treating reconstruction as a 3D plane detection problem. We use three definitions of true positive. ({\it All}) requires Mask IoU $\geq 0.5$, Normal error $\le 30^\circ$, and Offset error $\leq 1$m. ({\it -Offset}) removes the offset condition; ({\it -Normal}) removes the normal condition.}
    \label{tab:main}
        \begin{tabular}{@{~~}l@{~~}c@{~~}c@{~~}c@{~~}}
            \toprule
            Methods & All   & -Offset & -Normal \\
            \midrule
            Odometry~\cite{raposo2013plane} + PlaneRCNN~\cite{liu2019planercnn}           & 21.33 & 27.08 & 24.99 \\
            SuperGlue-GT Scale~\cite{sarlin2020superglue} + PlaneRCNN~\cite{liu2019planercnn}         & 30.06 & 33.24 & 33.52 \\
            LoFTR-DS-GT Scale~\cite{Sun2021} & 33.31 & 36.17 & 35.72 \\
            Camera Branch~\cite{Jin2021} + PlaneRCNN~\cite{liu2019planercnn}       &  29.44		& 35.25	&	31.67  \\
            Sparse Planes \cite{Jin2021}  (No Continuous) & 35.87 & 42.13 & 38.8 \\
            Proposed & \textbf{37.62} & \textbf{43.19} & \textbf{40.36} \\ \hline           
            Sparse Planes \cite{Jin2021} (Full) &  36.02	&	42.01	&	39.04 \\
            \bottomrule
        \end{tabular}
\end{table}

Our primary point of comparison is the wide baseline two view case. This two-view case has been extensively studied and benchmarked in~\cite{Jin2021}. We have shown qualitative results of the full system in Fig.~\ref{fig:results} by itself and show a comparison with Sparse Planes in Fig.~\ref{fig:recon_comparison}. We now discuss each aspect of performance.

\begin{figure}[t!]
	\centering
			{\includegraphics[width=\linewidth]{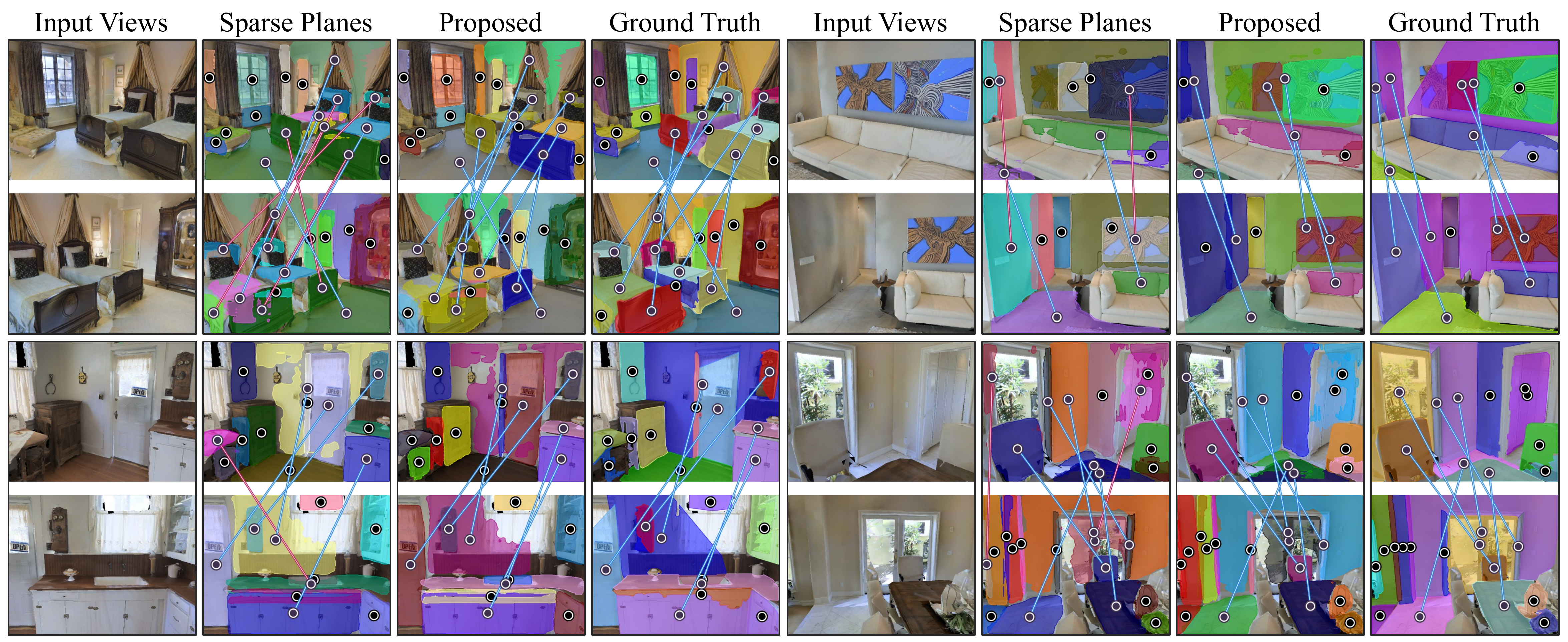}}
\caption{\textbf{Plane Comparison.} Matching surfaces across large view changes is challenging. Multiple surfaces may be similar in appearance, causing correspondence mixups like bed footboards (top left) or paintings (top right). By jointly refining planes across images via a transformer, the proposed method better associates across images. It can also reduce inconsistent outlier detections (bottom).}
	\label{fig:plane_comparison}
\end{figure}

\parnobf{Plane Correspondence Results} As reported in Table~\ref{tab:correspondence}, the PlaneFormer substantially increases IPAA across multiple metrics compared to Sparse Planes~\cite{Jin2021}. We show qualitative results on Fig.~\ref{fig:plane_comparison}, including one of the images that Jin \textit{et al.}~\cite{Jin2021} reported as a representative failure mode of their system. This particular case is challenging for geometry-based optimization since the bed footboards are co-planar and similar in appearance. These have similar appearance and plane parameter features; our mask token, however, can separate them out.

\parnobf{Relative Camera Pose Results} We next evaluate relative camera pose. Our results in Table~\ref{tab:camera} show that the PlaneFormer   outperforms most other approaches that do not do bundle adjustment on feature correspondence: by integrating plane information, the proposed system improves over the camera branch by over 10\% in translation accuracy and reduces rotation error by 22\% (relative). The approach outperforms the Sparse Planes system before continuous optimization. Even when Sparse Planes performs this step, our approach outperforms it in all but one metric. Our approach is competitive with SuperGlue~\cite{sarlin2020superglue} while LoFTR~\cite{Sun2021} outperforms competing systems in rotation estimation. Since these point-feature based approaches do not provide translation scale, we see them as complementary. Future systems might benefit from both points and planes.


\parnobf{Full Scene Evaluation Results} We finally report full scene evaluation results (AP) in Table~\ref{tab:main}. Our approach outperforms alternate methods, including  the full version of Sparse Planes. The relative performance gains of our method are smaller than compared to those for plane correspondence. However, it is important to note that the full scene evaluations is limited by every component. We hypothesize that one of the current key limiting factors is the accuracy of single-view reconstruction, or the initial PlaneRCNN~\cite{liu2019planercnn}.


\subsection{Wide Baseline Multiview Case}
\label{sec:experiments_multi}

\begin{figure}[t!]
	\centering
			{\includegraphics[width=\linewidth]{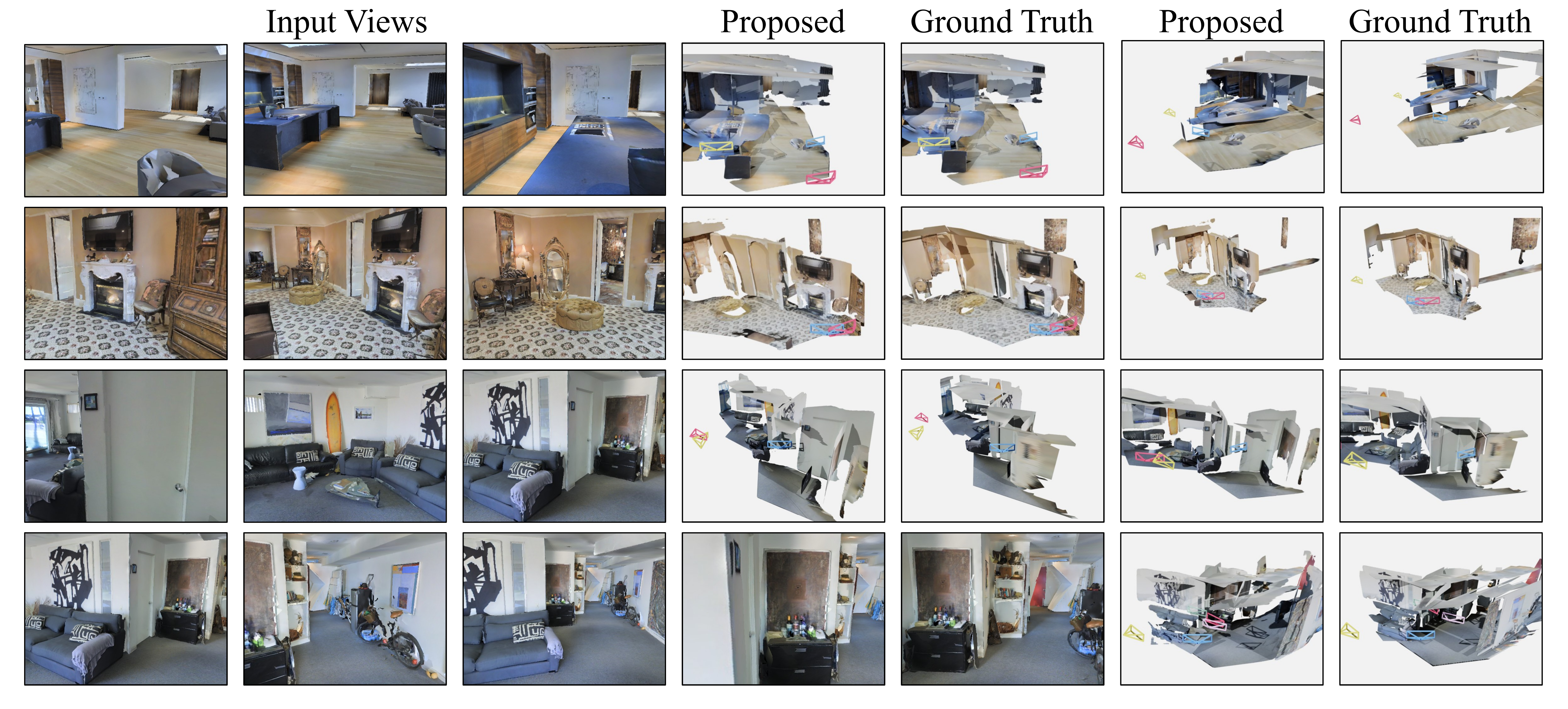}}
\caption{\textbf{Multiview Test Results.} With 3 views, our approach model can often construct extensive reconstruction of rooms (top 3 rows). With 5 views, the model continues to stitch larger sets of planes together effectively (bottom row).}
	\label{fig:recon_multiview}
\end{figure}

We next report the multiview case. The multiview case is substantially more challenging than performing pairwise reconstruction since the output must be a single coherent reconstruction. For instance, in relative camera pose, the composition of the rotation from image 1 to image 2 and the rotation from image 2 to image 3 must be the rotation from image 1 to image 3. 

\parnobf{Baselines} We extend the top baselines from the two view case to the multiview case. For fair comparison, we apply each baseline to the same view graph that is used for our method (defined in \S\ref{sec:method_inference}). We report both the full version of \cite{Jin2021} and the version without the continuous optimization ({\it No Cont.}~\cite{Jin2021}). For correspondence, we additionally report the appearance feature only baseline, which outperforms~\cite{Qian2020,cai2020messytable}. For camera pose estimation, we report the Camera Branch (Camera) of~\cite{Jin2021}, which outperforms multiple other baselines such as~\cite{raposo2013plane,Qian2020}.

\parnobf{Quantitative Results} We report correspondence results in Table~\ref{tab:multicorrespondence}. As was the case in the 2-view setting, our approach substantially outperforms the baselines. Overall performance reduces as the number of views increases; this is because the space covered often spreads out as the number of images increase, raising the difficulty of reconstruction. We next report relative camera pose estimation results in Table~\ref{tab:multicamera}. Trends are similar to the 2-view case: our method is competitive with the full pipeline of Jin et al.~\cite{Jin2021} in camera estimation, and often surpasses it. Our approach also substantially outperforms the top prediction from the Camera Branch and the discrete optimization version of~\cite{Jin2021}.

\parnobf{Qualitative Results} We show qualitative results on 3- and 5-view inputs in Figure~\ref{fig:recon_multiview}. Our method can often generate high quality scenes.


\begin{table}[t]
\centering
\caption{{\bf Multiview Evaluation: Plane Correspondence} We report IPAA-X for 3- and 5-view datasets. Our approach continues to substantially outperform baseline methods (but overall performance drops due to the increasing difficulty of the task).} \label{tab:multicorrespondence}
\begin{tabular}{lcccccccc} \toprule
 & ~~~ & \multicolumn{3}{c}{3-view IPAA-X} & ~~~ & \multicolumn{3}{c}{5-view IPAA-X}  \\ 
 &  & IPAA-100 & IPAA-90 & IPAA-80 & 
 & IPAA-100 & IPAA-90 & IPAA-80 \\ \midrule
Appearance & & 
5.94 & 20.28 & 52.97 & &
1.45 & 13.68 & 52.37 
\\
SparsePlanes~\cite{Jin2021} & &
9.95 & 23.77 & 51.16 & &
4.87 & 16.58 & 41.45 
\\
Proposed & &
\bf 14.60 & \bf 32.69 & \bf 66.15 & &
\bf 5.92 & \bf 20.66 & \bf 55.92 
\\ \bottomrule
\end{tabular}
\end{table}



\begin{table}[t]
\centering
\caption{{\bf Multiview Evaluation: Relative Camera Pose Estimation} We report the same metrics as the two view case, while running on the 3- and 5-view dataset.} \label{tab:multicamera}
\resizebox{1.0\linewidth}{!}{
\begin{tabular}{lcccccccccccc} \toprule
 & \multicolumn{6}{c}{3-view} & \multicolumn{6}{c}{5-view} \\
 &  \multicolumn{3}{c}{Transl. Error (m)} & \multicolumn{3}{c}{Rot. Error (deg)} 
 &  \multicolumn{3}{c}{Transl. Error (m)} & \multicolumn{3}{c}{Rot. Error (deg)} \\
 & Med. & Mean & $\le 1$m & 
 Med. & Mean & $\le 30^\circ$ 
 & Med. & Mean. & $\le 1$m & 
 Med. & Mean & $\le 30^\circ$  \\ \midrule
Camera~\cite{Jin2021} & 1.25 & 2.21 &  41.47 & 9.40 & 37.08 & 71.71 & 1.69 & 2.80 & 29.61 & 13.72 & 48.07 & 63.55 
\\
No Cont.~\cite{Jin2021} & 1.15 & 2.02 & 43.67 & 8.97 & 30.89 & \bf 75.97 & 1.62 & 2.73 & 31.58 & 12.08 & 44.99 & 64.08 \\
Proposed & \bf 0.83 & 1.81 & \bf 56.69 & \bf 7.88 & 32.22 & 74.94 & \bf 1.10 & 2.33 & 47.24 & \bf 9.52 & \bf 43.22 & \bf 67.5

\\ \midrule
Full~\cite{Jin2021} & 0.84 & \bf 1.74 & 54.91 & 8.83 & \bf 30.19 & 75.58 & 1.13 & \bf 2.29 & \bf 47.37 & 11.35 & 44.16 & 64.21
\\ \bottomrule
\end{tabular}}
\end{table}

\subsection{Ablations} 
\label{sec:experiments_ablation}

We finally analyze ablations of the method in Table \ref{tab:ablation}. We report the IPAA-90 and average camera pose translation and rotation errors. In all cases, ablations follow the same training procedure and are trained until validation accuracy plateaus. Full details and comparisons appear in supplement.

\parnobf{Feature ablations} To test feature importance, we report results when each feature has been removed from the token. For fair comparison, we keep transformer feature size equal to the full model by mapping inputs through an MLP. 
Table \ref{tab:ablation} (left) shows all three sets of features are important for performance. Removing appearance features causes the largest drop in plane correspondence, likely due to the importance of appearance when matching many planes across images. In contrast, removing plane parameters is most damaging to camera accuracy. As plane parameters represent position and orientation, this comparison indicates the position and orientation of planes are a powerful signal for inferring relative camera pose across images. Mask features have little impact on camera performance, but are still important for plane correspondence.

\parnobf{Network ablations} We next test the importance of the transformer and camera residual (Table~\ref{tab:ablation}, right). Our no transformer model simply applies the MLP heads for plane correspondence, camera correctness, and residuals. The plane features do not interact in this model, but it outperforms all prior baselines for plane correspondence and is competitive in camera pose, which illustrates the value of discriminatively learned correspondence rather than optimization. However, adding a transformer to enable inter-plane interaction substantially improve performance, with IPAA-90 increasing by nearly 8\%. 
The camera residual is also an important design decision, enabling refinement on predicted cameras. Note even without the residual, camera performance is similar to or better than all baselines. The camera residual does not impact plane predictions. 

\begin{table}[t]
    \caption{{\bf Ablations.}  We perform ablations of input features (left) and network design (right). We report IPAA-90 and relative camera pose translation and rotation error.}
    \label{tab:abl}
    
        \begin{minipage}[t]{.5\linewidth}
        \centering
            \begin{tabular}{lccc}
                \toprule
                Feature & Plane & Trans. & Rot. \\
                Ablation & IPAA-90 $\uparrow$ & Mean $\downarrow$ & Mean $\downarrow$ \\
                \midrule
                Proposed & \textbf{40.6} & \textbf{1.19} & 22.20 \\
                - Appearance & 26.9 & 1.23 & 22.78 \\
                - Plane & 35.2 & 1.32 & 25.92 \\
                - Mask & 34.5 & 1.26 & \textbf{21.21} \\
                \bottomrule
            \end{tabular}
        \end{minipage} 
        \begin{minipage}[t]{.5\linewidth}
        \centering
            \vspace*{-0.565in}
            \begin{tabular}{lccc}
                \toprule
                Network & Plane & Trans. & Rot. \\
                Ablation & IPAA-90 $\uparrow$ & Mean $\downarrow$ & Mean $\downarrow$ \\
                \midrule
                Proposed & \textbf{40.6} & \textbf{1.19} & \textbf{22.20} \\
                - Transformer & 32.7 & 1.48 & 26.43 \\
                - Residual & 40.6 & 1.34 & 22.38 \\
                \bottomrule
            \end{tabular}
        \end{minipage}
        
    \label{tab:ablation}
    \vspace{-0.1in}
\end{table}

\section{Conclusion}

We have introduced a new model for performing reconstructions between images separated by wide baselines. Our approach replaces hand-designed optimization with a discriminatively learned transformer and shows substantial improvements over the state of the art across multiple metrics and settings. 

\vspace{1mm}
\parnobf{Acknowledgements} This work was supported by the
DARPA Machine Common Sense Program. We would like to thank Richard Higgins and members of the Fouhey lab for helpful discussions and feedback.

\clearpage
%
%
\bibliographystyle{splncs04}
\bibliography{egbib}

\begin{thebibliography}{10}
\providecommand{\url}[1]{\texttt{#1}}
\providecommand{\urlprefix}{URL }
\providecommand{\doi}[1]{https://doi.org/#1}

\bibitem{agarwal2010bundle}
Agarwal, S., Snavely, N., Seitz, S.M., Szeliski, R.: Bundle adjustment in the
  large. In: ECCV (2010)

\bibitem{bloem_2019}
Bloem, P.: Transformers from scratch (2019),
  \url{http://peterbloem.nl/blog/transformers}

\bibitem{bozic2021transformerfusion}
Bozic, A., Palafox, P., Thies, J., Dai, A., Nie{\ss}ner, M.: Transformerfusion:
  Monocular rgb scene reconstruction using transformers. NeurIPS  (2021)

\bibitem{Cai2021Extreme}
Cai, R., Hariharan, B., Snavely, N., Averbuch-Elor, H.: Extreme rotation
  estimation using dense correlation volumes. In: CVPR (2021)

\bibitem{cai2020messytable}
Cai, Z., Zhang, J., Ren, D., Yu, C., Zhao, H., Yi, S., Yeo, C.K., Loy, C.C.:
  Messytable: Instance association in multiple camera views. In: ECCV (2020)

\bibitem{chang2017matterport3d}
Chang, A., Dai, A., Funkhouser, T., Halber, M., Niessner, M., Savva, M., Song,
  S., Zeng, A., Zhang, Y.: Matterport3d: Learning from rgb-d data in indoor
  environments. In: 3DV (2017)

\bibitem{chen2021mvsnerf}
Chen, A., Xu, Z., Zhao, F., Zhang, X., Xiang, F., Yu, J., Su, H.: Mvsnerf: Fast
  generalizable radiance field reconstruction from multi-view stereo. In: ICCV
  (2021)

\bibitem{Chen_2021_CVPR}
Chen, K., Snavely, N., Makadia, A.: Wide-baseline relative camera pose
  estimation with directional learning. In: CVPR (2021)

\bibitem{chen2020oasis}
Chen, W., Qian, S., Fan, D., Kojima, N., Hamilton, M., Deng, J.: Oasis: A
  large-scale dataset for single image 3d in the wild. In: CVPR (2020)

\bibitem{choy2020deep}
Choy, C., Dong, W., Koltun, V.: Deep global registration. In: CVPR (2020)

\bibitem{choy20163d}
Choy, C.B., Xu, D., Gwak, J., Chen, K., Savarese, S.: 3d-r2n2: A unified
  approach for single and multi-view 3d object reconstruction. In: ECCV (2016)

\bibitem{eigen15}
Eigen, D., Fergus, R.: Predicting depth, surface normals and semantic labels
  with a common multi-scale convolutional architecture. In: ICCV (2015)

\bibitem{banani2020unsupervisedrr}
El~Banani, M., Gao, L., Johnson, J.: Unsupervised r\&r: Unsupervised point
  cloud registration via differentiable rendering. In: CVPR (2021)

\bibitem{fan2017point}
Fan, H., Su, H., Guibas, L.J.: A point set generation network for 3d object
  reconstruction from a single image. In: CVPR (2017)

\bibitem{furukawa2009manhattan}
Furukawa, Y., Curless, B., Seitz, S.M., Szeliski, R.: Manhattan-world stereo.
  In: CVPR (2009)

\bibitem{gkioxari2019mesh}
Gkioxari, G., Malik, J., Johnson, J.: Mesh r-cnn. In: ICCV (2019)

\bibitem{Hartley04}
Hartley, R.I., Zisserman, A.: Multiple View Geometry in Computer Vision.
  Cambridge University Press, ISBN: 0521540518 (2004)

\bibitem{he2017mask}
He, K., Gkioxari, G., Doll{\'a}r, P., Girshick, R.: Mask r-cnn. In: ICCV (2017)

\bibitem{hoiem2005geometric}
Hoiem, D., Efros, A.A., Hebert, M.: Geometric context from a single image. In:
  ICCV (2005)

\bibitem{huang2018deepmvs}
Huang, P.H., Matzen, K., Kopf, J., Ahuja, N., Huang, J.B.: Deepmvs: Learning
  multi-view stereopsis. In: CVPR (2018)

\bibitem{huang2018deep}
Huang, Z., Li, T., Chen, W., Zhao, Y., Xing, J., LeGendre, C., Luo, L., Ma, C.,
  Li, H.: Deep volumetric video from very sparse multi-view performance
  capture. In: ECCV (2018)

\bibitem{jain2021putting}
Jain, A., Tancik, M., Abbeel, P.: Putting nerf on a diet: Semantically
  consistent few-shot view synthesis. In: ICCV (2021)

\bibitem{jiang2020local}
Jiang, C., Sud, A., Makadia, A., Huang, J., Nie{\ss}ner, M., Funkhouser, T.,
  et~al.: Local implicit grid representations for 3d scenes. In: CVPR (2020)

\bibitem{Jin2021}
Jin, L., Qian, S., Owens, A., Fouhey, D.F.: Planar surface reconstruction from
  sparse views. In: ICCV (2021)

\bibitem{Jin2020}
Jin, Y., Mishkin, D., Mishchuk, A., Matas, J., Fua, P., Yi, K.M., Trulls, E.:
  {Image Matching across Wide Baselines: From Paper to Practice}. IJCV  (2020)

\bibitem{kar2017learning}
Kar, A., H{\"a}ne, C., Malik, J.: Learning a multi-view stereo machine. In:
  NeurIPS (2017)

\bibitem{kopf2021robust}
Kopf, J., Rong, X., Huang, J.B.: Robust consistent video depth estimation. In:
  CVPR (2021)

\bibitem{kuhn1955hungarian}
Kuhn, H.W.: The hungarian method for the assignment problem. Naval Research
  Logistics Quarterly  (1955)

\bibitem{li2018megadepth}
Li, Z., Snavely, N.: Megadepth: Learning single-view depth prediction from
  internet photos. In: CVPR (2018)

\bibitem{lin2021barf}
Lin, C.H., Ma, W.C., Torralba, A., Lucey, S.: Barf: Bundle-adjusting neural
  radiance fields. In: ICCV (2021)

\bibitem{lin2021metro}
Lin, K., Wang, L., Liu, Z.: End-to-end human pose and mesh reconstruction with
  transformers. In: CVPR (2021)

\bibitem{lin2017feature}
Lin, T.Y., Doll{\'a}r, P., Girshick, R., He, K., Hariharan, B., Belongie, S.:
  Feature pyramid networks for object detection. In: CVPR (2017)

\bibitem{lindenberger2021pixel}
Lindenberger, P., Sarlin, P.E., Larsson, V., Pollefeys, M.: Pixel-perfect
  structure-from-motion with featuremetric refinement. In: ICCV (2021)

\bibitem{liu2019planercnn}
Liu, C., Kim, K., Gu, J., Furukawa, Y., Kautz, J.: Planercnn: 3d plane
  detection and reconstruction from a single image. In: CVPR (2019)

\bibitem{liu2018planenet}
Liu, C., Yang, J., Ceylan, D., Yumer, E., Furukawa, Y.: Planenet: Piece-wise
  planar reconstruction from a single rgb image. In: CVPR (2018)

\bibitem{lowe2004distinctive}
Lowe, D.G.: Distinctive image features from scale-invariant keypoints. IJCV
  (2004)

\bibitem{ma2004invitation}
Ma, Y., Soatto, S., Ko{\v{s}}eck{\'a}, J., Sastry, S.: An invitation to 3-d
  vision: from images to geometric models. Springer (2004)

\bibitem{mescheder2019occupancy}
Mescheder, L., Oechsle, M., Niemeyer, M., Nowozin, S., Geiger, A.: Occupancy
  networks: Learning 3d reconstruction in function space. In: CVPR (2019)

\bibitem{mildenhall2020nerf}
Mildenhall, B., Srinivasan, P.P., Tancik, M., Barron, J.T., Ramamoorthi, R.,
  Ng, R.: Nerf: Representing scenes as neural radiance fields for view
  synthesis. In: ECCV (2020)

\bibitem{mur2015orb}
Mur-Artal, R., Montiel, J.M.M., Tardos, J.D.: Orb-slam: a versatile and
  accurate monocular slam system. TOG  (2015)

\bibitem{Pritchett98a}
Pritchett, P., Zisserman, A.: Wide baseline stereo matching. In: ICCV (1998)

\bibitem{Qian2020}
Qian, S., Jin, L., Fouhey, D.F.: Associative3d: Volumetric reconstruction from
  sparse views. In: ECCV (2020)

\bibitem{Ranftl2020}
Ranftl, R., Lasinger, K., Hafner, D., Schindler, K., Koltun, V.: Towards robust
  monocular depth estimation: Mixing datasets for zero-shot cross-dataset
  transfer. TPAMI  (2020)

\bibitem{raposo2013plane}
Raposo, C., Louren{\c{c}}o, M., Antunes, M., Barreto, J.P.: Plane-based
  odometry using an rgb-d camera. In: BMVC (2013)

\bibitem{sarlin2020superglue}
Sarlin, P.E., DeTone, D., Malisiewicz, T., Rabinovich, A.: Superglue: Learning
  feature matching with graph neural networks. In: CVPR (2020)

\bibitem{schonberger2016structure}
Schonberger, J.L., Frahm, J.M.: Structure-from-motion revisited. In: CVPR
  (2016)

\bibitem{schoenberger2016mvs}
Sch\"{o}nberger, J.L., Zheng, E., Pollefeys, M., Frahm, J.M.: {Pixelwise View
  Selection for Unstructured Multi-View Stereo}. In: ECCV (2016)

\bibitem{song2017semantic}
Song, S., Yu, F., Zeng, A., Chang, A.X., Savva, M., Funkhouser, T.: Semantic
  scene completion from a single depth image. In: CVPR (2017)

\bibitem{Sun2021}
Sun, J., Shen, Z., Wang, Y., Bao, H., Zhou, X.: {LoFTR}: Detector-free local
  feature matching with transformers. CVPR  (2021)

\bibitem{sun2021neuralrecon}
Sun, J., Xie, Y., Chen, L., Zhou, X., Bao, H.: Neuralrecon: Real-time coherent
  3d reconstruction from monocular video. In: CVPR (2021)

\bibitem{teed2021droid}
Teed, Z., Deng, J.: Droid-slam: Deep visual slam for monocular, stereo, and
  rgb-d cameras. NeurIPS  (2021)

\bibitem{triggs1999bundle}
Triggs, B., McLauchlan, P.F., Hartley, R.I., Fitzgibbon, A.W.: Bundle
  adjustment—a modern synthesis. In: International workshop on vision
  algorithms (1999)

\bibitem{ummenhofer2017demon}
Ummenhofer, B., Zhou, H., Uhrig, J., Mayer, N., Ilg, E., Dosovitskiy, A., Brox,
  T.: Demon: Depth and motion network for learning monocular stereo. In: CVPR
  (2017)

\bibitem{vaswani2017attention}
Vaswani, A., Shazeer, N., Parmar, N., Uszkoreit, J., Jones, L., Gomez, A.N.,
  Kaiser, {\L}., Polosukhin, I.: Attention is all you need. In: NeurIPS (2017)

\bibitem{wang2018pixel2mesh}
Wang, N., Zhang, Y., Li, Z., Fu, Y., Liu, W., Jiang, Y.G.: Pixel2mesh:
  Generating 3d mesh models from single rgb images. In: ECCV (2018)

\bibitem{wang2021ibrnet}
Wang, Q., Wang, Z., Genova, K., Srinivasan, P., Zhou, H., Barron, J.T.,
  Martin-Brualla, R., Snavely, N., Funkhouser, T.: Ibrnet: Learning multi-view
  image-based rendering. In: CVPR (2021)

\bibitem{wang2020tartanvo}
Wang, W., Hu, Y., Scherer, S.: Tartanvo: A generalizable learning-based vo. In:
  CoRL (2020)

\bibitem{Wang15}
Wang, X., Fouhey, D.F., Gupta, A.: Designing deep networks for surface normal
  estimation. In: CVPR (2015)

\bibitem{wiles2020synsin}
Wiles, O., Gkioxari, G., Szeliski, R., Johnson, J.: Synsin: End-to-end view
  synthesis from a single image. In: CVPR (2020)

\bibitem{planeformer}
Wong, S.: Takaratomy transformers henkei octane,
  \url{https://live.staticflickr.com/3166/2970928056_c3b59be5ca_b.jpg}

\bibitem{wu20083d}
Wu, C., Clipp, B., Li, X., Frahm, J.M., Pollefeys, M.: 3d model matching with
  viewpoint-invariant patches (vip). In: CVPR (2008)

\bibitem{yang2018recovering}
Yang, F., Zhou, Z.: Recovering 3d planes from a single image via convolutional
  neural networks. In: ECCV (2018)

\bibitem{yi2018learning}
Yi, K.M., Trulls, E., Ono, Y., Lepetit, V., Salzmann, M., Fua, P.: Learning to
  find good correspondences. In: CVPR (2018)

\bibitem{yu2020pixelnerf}
Yu, A., Ye, V., Tancik, M., Kanazawa, A.: pixelnerf: Neural radiance fields
  from one or few images. In: CVPR (2021)

\bibitem{yu2019single}
Yu, Z., Zheng, J., Lian, D., Zhou, Z., Gao, S.: Single-image piece-wise planar
  3d reconstruction via associative embedding. In: CVPR (2019)

\bibitem{zhang2019learning}
Zhang, J., Sun, D., Luo, Z., Yao, A., Zhou, L., Shen, T., Chen, Y., Quan, L.,
  Liao, H.: Learning two-view correspondences and geometry using order-aware
  network. In: ICCV (2019)

\bibitem{zhang1994iterative}
Zhang, Z.: Iterative point matching for registration of free-form curves and
  surfaces. IJCV  (1994)

\bibitem{zhang2021consistent}
Zhang, Z., Cole, F., Tucker, R., Freeman, W.T., Dekel, T.: Consistent depth of
  moving objects in video. TOG  (2021)

\bibitem{zhao2021point}
Zhao, H., Jiang, L., Jia, J., Torr, P.H., Koltun, V.: Point transformer. In:
  ICCV (2021)

\end{thebibliography}
\clearpage

\appendix
\title{Supplementary Material for PlaneFormers} 


\titlerunning{PlaneFormers}
%

\index{Fouhey, David F.}
\author{Samir Agarwala \and
Linyi Jin \and Chris Rockwell \and David F. Fouhey} 
\authorrunning{Agarwala et al.}
%
\institute{University of Michigan, Ann Arbor\\
\email{\{samirag,jinlinyi,cnris,fouhey\}@umich.edu}}
\maketitle

The supplemental material consists of both video results and this PDF. The PDF portion of the supplemental material shows: detailed descriptions of model architectures (Section 1), details about the experimental setup (Section 2), and additional results (Section 3). Video results visualize qualitative reconstructions from the main paper from a variety of viewpoints.
\section{Model Architecture}

\begin{table}[h!]
\caption{\textbf{Model Architecture}. We define the number of planes from view $i$ be $M_i$, $M =  M_1 + M_2$, dimension $D = 899$. Embeddings are passed through a 5-layer transformer encoder which has 1 head, dropout probability of 0.1 and a feedforward network dimension of 2048. We create a pair-wise feature tensor of dimension $M \times M \times 4D$ and pass this tensor through 4 separate MLP heads to estimate the plane correspondence, camera correspondence, rotation residual and translation residual. We mask out entries in the MLP outputs such that only the pairwise predictions between planes across views are considered during average pooling in the camera correspondence and residual heads (note: the shape after masking in the table represents non-zero entries). Finally, we apply a sigmoid function to the plane correspondence and the camera correspondence scores, and extract plane correspondences across views.}\label{tab:model_arch}
\centering
\scriptsize
\resizebox{0.75\textwidth}{!}{
\begin{tabular}{@{}lllc@{}}
\toprule
Index & Inputs & Operation                                                                                                                                                                                                                                                                                                                                                & Output Shape           \\ \midrule
(1)   & Inputs & Input Embedding                                                                                                                                                                                                                                                                                                                                          & $M \times D$           \\
\hline
(2)   & (1)    & 5-Layer Transformer Encoder                                                                                                                                                                                                                                                                                                                              & $M \times D$           \\
\hline
(3)   & (2)    & Create Pair-wise Feature Tensor                                                                                                                                                                                                                                                                                                                          & $M \times M \times 4D$ \\
\hline
(4)   & (3)    & \begin{tabular}[c]{@{}l@{}}Plane Correspondence: Linear($4D \rightarrow 2D$), \\ Linear($2D \rightarrow D$), Linear($D \rightarrow D/2$), \\ Linear($D/2 \rightarrow D/4$), Linear($D/4  \rightarrow 1$),\\ Sigmoid($M \times M$), \\
Extract Submatrix($M \times M \rightarrow M_1 \times M_2$)\end{tabular}                                                                      & $M_1 \times M_2$       \\
\hline
(5)   & (3)    & \begin{tabular}[c]{@{}l@{}}Camera Correspondence: Linear($4D \rightarrow 2D$), \\ Linear($2D \rightarrow D$), Linear($D \rightarrow D/2$), \\ Linear($D/2 \rightarrow D/4$), Linear($D/4  \rightarrow 1$),\\ Mask Matrix($M \times M \rightarrow M_1 \times M_2$),  \\ AveragePool($M_1 \times M_2 \rightarrow 1$), Sigmoid(1)\end{tabular}                           & 1                      \\
\hline
(6)   & (3)    & \begin{tabular}[c]{@{}l@{}}Rotation Residual: Linear($4D \rightarrow 2D$), \\ Linear($2D \rightarrow D$), Linear($D \rightarrow D/2$), \\ Linear($D/2 \rightarrow D/4$), Linear($D/4  \rightarrow 4$),\\ Mask Matrix($M \times M \times 4 \rightarrow M_1 \times M_2 \times 4$),  \\ AveragePool($M_1 \times M_2 \times 4 \rightarrow 4$)\end{tabular}    & 4                      \\
\hline
(7)   & (3)    & \begin{tabular}[c]{@{}l@{}}Translation Residual: Linear($4D \rightarrow 2D$), \\ Linear($2D \rightarrow D$), Linear($D \rightarrow D/2$), \\ Linear($D/2 \rightarrow D/4$), Linear($D/4  \rightarrow 3$),\\ Mask Matrix($M \times M \times 3 \rightarrow M_1 \times M_2 \times 3$),  \\ AveragePool($M_1 \times M_2 \times 3 \rightarrow 3$)\end{tabular} & 3                      \\ \bottomrule
\end{tabular}
    }
\end{table}

\section{Experimental Details}
\subsection{Multiview Dataset Creation}
The two view dataset is the same as~\cite{Jin2021}. For 3-view and 5-view datasets, we use the single images sampled by~\cite{Jin2021}, then randomly sample combinations of images within each floor of the house. We select sets of images where each image in any pair has $\geq$ 3 matches and $\geq$ 3 unique planes. The maximum number of sets per floor is 10. We finally get a three-view test set of size 258 and a five-view test set of size 76. We do not need training set or validation set for 3-view and 5-view cases since our network is not trained on the multiview dataset. 

\subsection{Multiview Evaluation}
We compared our proposed approach with baselines on the same view graph that was built as discussed in the approach section. For multi-view evaluation, we consider all combinations of input views in a sample and independently compute the pair-wise IPAA, rotation error and translation error for each combination. For instance, in the 3-view case, we consider the IPAA and camera error metrics independently between view 1 and view 2, view 1 and view 3, and view 2 and view 3. 

The relative camera transformations and plane correspondences between any combination of views is computed by chaining together relative camera transformations and plane correspondences across the created view graph, and is then compared to the ground-truth. Finally, we compute IPAA-X and camera error statistics over all combinations of views across samples in the test set (i.e. we consider each combination of views in a test sample as an independent datapoint while computing our metrics).

\subsection{Ablation Details}
For fair comparison, we use our same training setup for ablations. Ablations are trained until validation accuracy plateaus, which in practice is 40k iterations; the same as for the full model. 

\parnobf{Feature Ablations} Ablating features results in a smaller input feature space. For fair comparison, we therefore use a linear layer to project this smaller input feature to features the same size as the full model. Transformer layers then operate at the same size as in the case of the full model.

\parnobf{Model Ablations} In our full model, we classify
camera pose into clusters from \cite{Jin2021}, and predict a corrective residual camera pose to the predicted cluster. In the \textit{without residual} ablation, we remove this residual camera pose; this tests if the corrective residual improves predicted pose. Our \textit{without transformer} ablation takes input features as direct input to final camera and plane MLP layers, testing if the transformer improves plane features for final prediction.

\section{Additional Results}
\subsection{Quantitative Results}
\subsubsection{Additional Plane Correspondence Results}
We report IPAA-100 and IPAA-80 for the feature and network ablations in addition to IPAA-90 that was provided in our experiments in Tables \ref{tab:ft_abl_all_ipaa} and \ref{tab:net_abl_all_ipaa}, respectively.

\begin{table}
\caption{IPAA-100, IPAA-90 and IPAA-80 for feature ablations.}\label{tab:ft_abl_all_ipaa}
\centering
\begin{tabular}{lccc}
\toprule
 Feature            & \multicolumn{1}{l}{IPAA-100 ↑} & \multicolumn{1}{l}{IPAA-90 ↑} & \multicolumn{1}{l}{IPAA-80 ↑} \\ \midrule
Proposed     & \textbf{19.6}                  & \textbf{40.6}                 & \textbf{71.0}                 \\
- Appearance & 11.3                           & 26.9                          & 57.1                          \\
- Plane      & 16.5                           & 35.2                          & 65.6                          \\
- Mask       & 15.1                           & 34.5                          & 67.2                          \\ \bottomrule
\end{tabular}
\end{table}

\begin{table}
\caption{IPAA-100, IPAA-90 and IPAA-80 for network ablations. The IPAA-X results for the model without residual are the same as the proposed method since the camera residual affects the relative camera pose prediction but not the plane correspondences.}\label{tab:net_abl_all_ipaa}
\centering
\begin{tabular}{lccc}
\toprule
 Network             & \multicolumn{1}{l}{IPAA-100 ↑} & \multicolumn{1}{l}{IPAA-90 ↑} & \multicolumn{1}{l}{IPAA-80 ↑} \\ \midrule
Proposed      & \textbf{19.6}                  & \textbf{40.6}                 & \textbf{71.0}                 \\
- Transformer & 13.8                           & 32.7                          & 64.3                          \\
- Residual    & 19.6                           & 40.6                          & 71.0                          \\ \bottomrule
\end{tabular}
\end{table}

\subsubsection{Additional Relative Camera Pose Estimation Results} We report median error and $\%$ error {$\leq$} 1m or $30^{\circ}$ for the predicted relative translation and rotation for the feature and network ablations in addition to mean error that was provided in our experiments in Tables \ref{tab:camera_ft_abl_all} and \ref{tab:camera_net_abl_all} respectively.

\begin{table}[t]    
    \centering
    \caption{Median error, mean error and $\%$ error {$\leq$} 1m or $30^{\circ}$ for translation and rotation for the feature ablations.
    }
    \label{tab:camera_ft_abl_all}
        \begin{tabular}{@{\hskip5pt}l@{~}c@{~}c@{~}c@{~~~~~~}c@{~}c@{~}c@{\hskip5pt}}
        \toprule
        & \multicolumn{3}{c}{Translation} & \multicolumn{3}{c}{Rotation} \\
        Method & Med. & Mean & (${\leq}1$m) & Med. & Mean & (${\leq}30^{\circ}$) \\
        \midrule
        Proposed & \textbf{0.66} & \textbf{1.19} & \textbf{66.8} & 5.96 & 22.20 & 83.8 \\
        - Appearance & 0.69 & 1.23 & 65.2 & 6.01 & 22.78 & 83.3 \\
        - Plane & 0.81 &	1.32 &	59.6 &	10.34 &	25.92 &	81.4 \\
        - Mask & 0.75 &	1.26 &	61.9 &	\textbf{4.65} &	\textbf{21.21} &	\textbf{84.3} \\
        \bottomrule
        \end{tabular}
\end{table}

\begin{table}[t]    
    \centering
    \caption{Median error, mean error and $\%$ error {$\leq$} 1m or $30^{\circ}$ for translation and rotation for the network ablations.
    }
    \label{tab:camera_net_abl_all}
        \begin{tabular}{@{\hskip5pt}l@{~}c@{~}c@{~}c@{~~~~~~}c@{~}c@{~}c@{\hskip5pt}}
        \toprule
        & \multicolumn{3}{c}{Translation} & \multicolumn{3}{c}{Rotation} \\
        Method & Med. & Mean & (${\leq}1$m) & Med. & Mean & (${\leq}30^{\circ}$) \\
        \midrule
        Proposed & \textbf{0.66} & \textbf{1.19} & \textbf{66.8} & \textbf{5.96} & \textbf{22.20} & \textbf{83.8} \\
        - Transformer & 1.02 &	1.48 &	49 &	10.54 &	26.43 &	80.8 \\
        - Residual & 0.88 &	1.34 &	57.7 &	6.22 &	22.38 &	83.7 \\
        \bottomrule
        \end{tabular}
\end{table}

\subsection{Qualitative Results}

\parnobf{Video Results} Video results bring to life the reconstructions from the main paper. As stated, PlaneFormer planar reconstructions are often quite close to the ground truth even faced with large view changes and challenging coplanar settings. If 3 or 5 views are available, the model continues to produce coherent results.

\parnobf{Additional Examples} We include additional outputs for 2 input views (Figure \ref{fig:supp_2view}), and 3 and 5 views (Figure \ref{fig:supp_3_5view}). These results are consistent with those in the paper: plane correspondences tend to be accurate even in challenging cases, and reconstructions are reasonable in very large view change cases and accurate in smaller view change cases.

\begin{figure}[t!]
	\centering
			{\includegraphics[width=\linewidth]{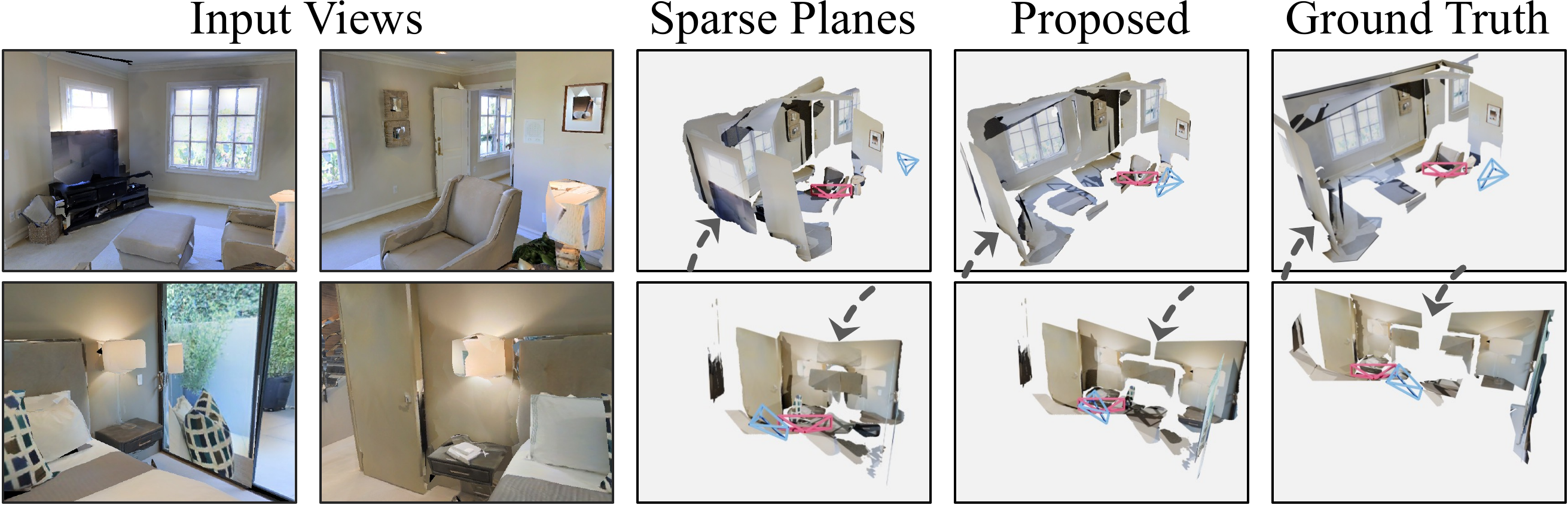}}
\caption{\textbf{Additional reconstruction comparison, extending Fig. 4} Sparse Plane reconstructions are a good baseline, but PlaneFormer yields superior results. It produces both better stitched planes (top), and more accurate camera (bottom).}
	\label{fig:supp:recon_comparison}
\end{figure}

\begin{figure}[t!]
	\centering
			{\includegraphics[width=\linewidth]{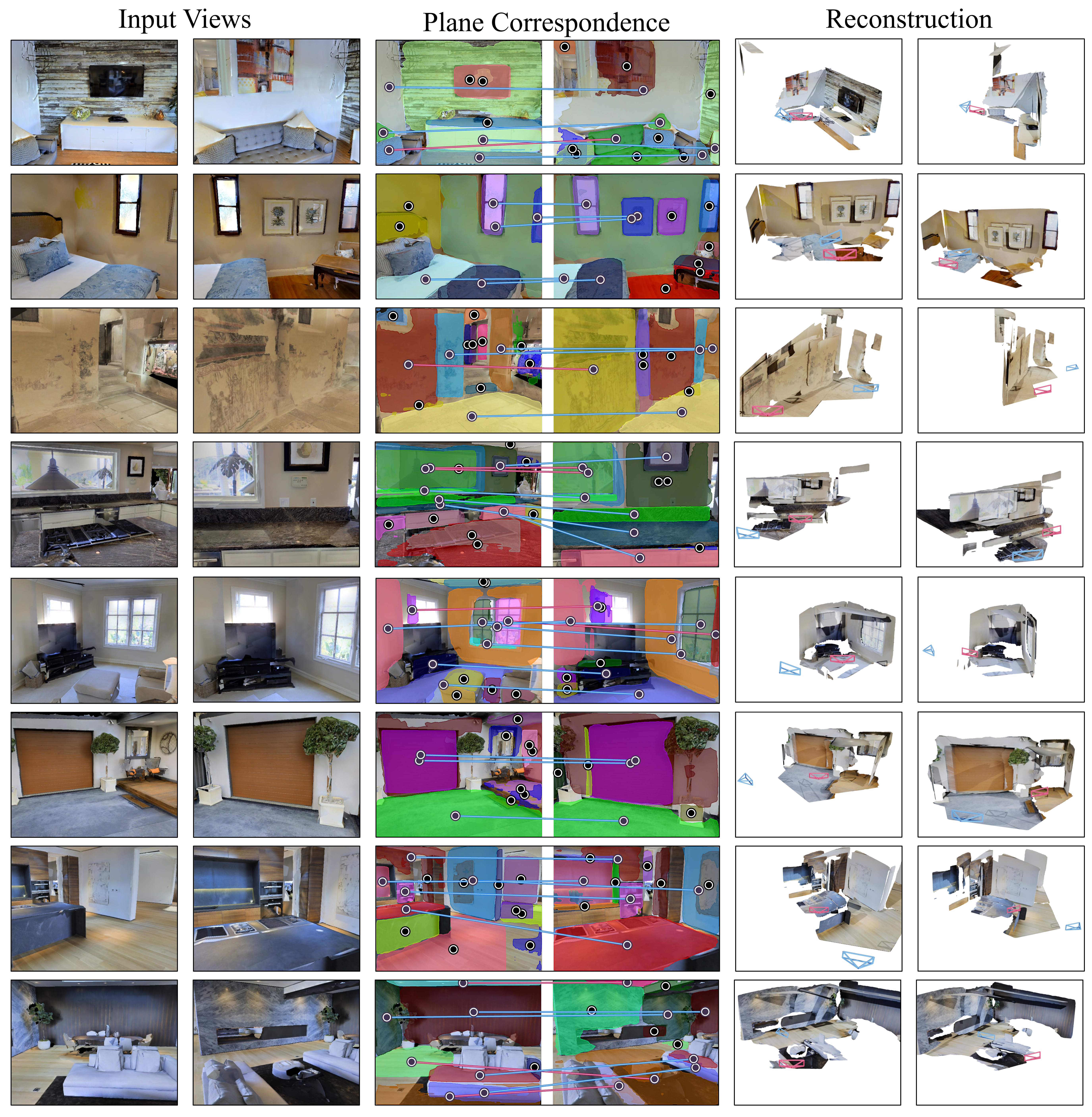}}
\caption{\textbf{Additional 2 View Results.}}
	\label{fig:supp_2view}
\end{figure}

\begin{figure}[t!]
	\centering
			{\includegraphics[width=\linewidth]{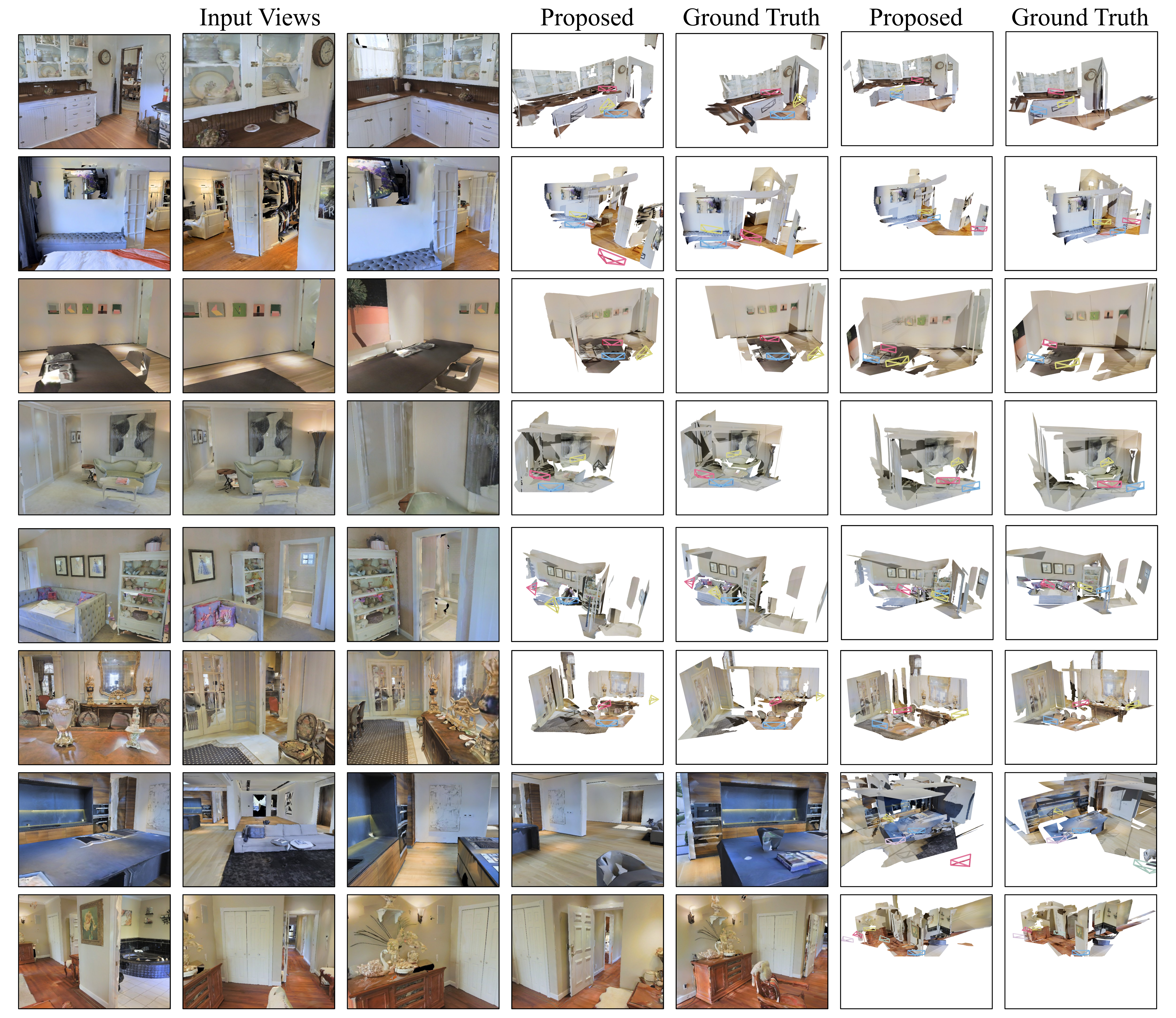}}
\caption{\textbf{Additional 3 and 5 View Results.}}
	\label{fig:supp_3_5view}
\end{figure}

\parnobf{Additional reconstruction comparison, extending Fig. 4} See Figure ~\ref{fig:supp:recon_comparison}. 

\begin{figure}[t!]
	\centering
			{\includegraphics[width=\linewidth]{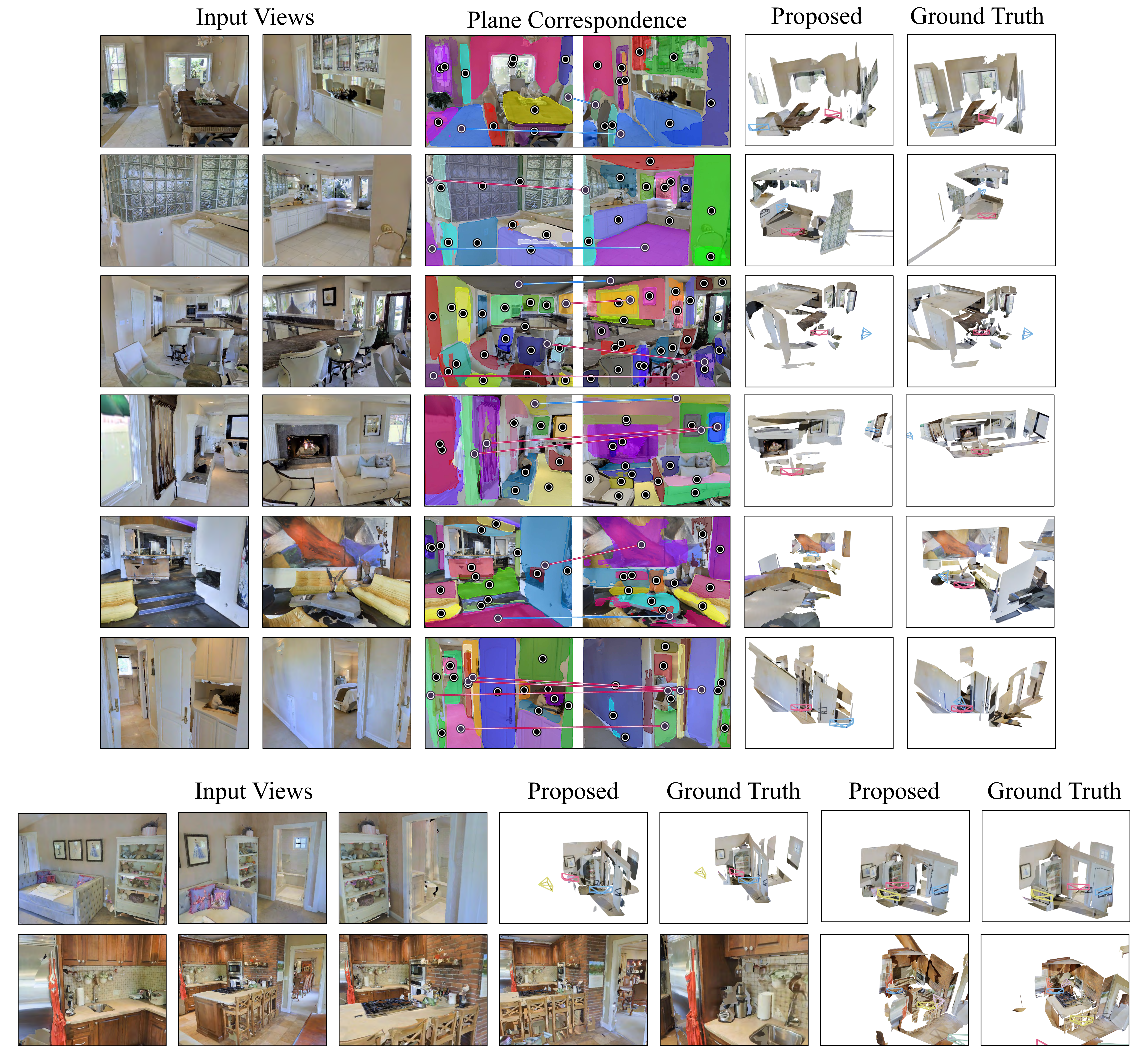}}
\caption{\textbf{Limitations and Failure Cases.}}
	\label{fig:supp_failure}
\end{figure}

\parnobf{Limitations and Failure Cases} We also include limitations and failure cases in Figure \ref{fig:supp_failure}. One limitation of a plane representation is that planes struggle to model small details in scenes, which sometimes leads to incomplete reconstructions (top two examples). The model may also perform poorly in some circumstances. Plane correspondences struggle when many small, similar objects are visible across large view change (second two examples). Predicting camera can sometimes be difficult given large view change leading to significant difference in appearance (bottom two examples in two-view case). Sometimes both camera and correspondence are poor (bottom example, two-view case). When more views are present, planes are not always fused cleanly, leading to intersections (final two examples, multi-view case). 

\clearpage
%
%

\end{document}